\documentclass[letterpaper, 10 pt, conference]{ieeeconf}

\IEEEoverridecommandlockouts

\usepackage{graphics} %
\usepackage{graphicx}
\usepackage{epsfig} %
\usepackage{amsmath} %
\usepackage{amssymb}  %
\usepackage{subfigure}
\usepackage{float}
\usepackage{url}
\usepackage{multirow}
\usepackage{subfigure}
\usepackage{wrapfig}
\usepackage{tabularx}
\usepackage{hyperref} %
\usepackage{color}
\usepackage{bm}
\usepackage{array}
\usepackage{booktabs}
\usepackage{algorithm,algorithmic}
\usepackage{siunitx}

\usepackage{gensymb}

\graphicspath{{./figs/}}

\newcolumntype{C}[1]{>{\centering}m{#1}}
\newcolumntype{L}{>{\centering\arraybackslash}m{0.7cm}}
\newcommand{\doctitle}{Imminent Collision Mitigation with Reinforcement Learning and Vision}
\newcommand{\docsubtitle}{}

\title{\LARGE \bf\doctitle\docsubtitle}

\author{Horia Porav and Paul Newman \thanks{{Authors are from the Oxford Robotics Institute, Dept. Engineering Science, University of Oxford, UK. \{\texttt{horia}, \texttt{pnewman}\}\texttt{@robots.ox.ac.uk}}}
}

\begin{document}
\maketitle
\thispagestyle{empty}
\pagestyle{empty}

\begin{abstract}

This work examines the role of reinforcement learning in reducing the severity of on-road collisions by controlling velocity and steering in situations in which contact is imminent. We construct a model, given camera images as input, that is capable of learning and predicting the dynamics of obstacles, cars and pedestrians, and train our policy using this model. Two policies that control both braking and steering are compared against a baseline where the only action taken is (conventional) braking in a straight line. The two policies are trained using two distinct reward structures, one where any and all collisions incur a fixed penalty, and a second one where the penalty is calculated based on already established delta-v models of injury severity. The results show that both policies exceed the performance of the baseline, with the policy trained using injury models having the highest performance.

\end{abstract}

\section{Introduction}\label{sec:introduction}
Both in the case of human-driven vehicles and autonomous vehicles, ensuring the safety of all traffic participants is paramount and often \emph{the} prime motivator in autonomous vehicle research. Part of this endeavour is considering what to do if a situation arose in which some sort of collision was inevitable. We need not here exhaustively contemplate the many conceivable causes of such a situation - it could be entirely due to uncontrollable third-party decision or an ``act-of-god'' event. However, we ought to consider what appropriate actions can be taken in the moments after it becomes apparent that a collision is likely or inevitable and before impact occurs. There is potential to reduce the total harm and injuries sustained during a collision scenario, especially in the case where emergency braking systems by themselves are unable to prevent a collision \cite{Chae2017}.

Current approaches can be grouped under two major categories:
\begin{itemize}
    \item emergency systems that only brake, without any steering input\cite{Kaempchen2009},\cite{Chae2017}
    \item emergency systems that are capable of both braking and steering, usually in the form of lane changing\cite{Shiller1998},\cite{Choi2012}
\end{itemize}

The major drawback of braking-only systems is that they cannot avoid collisions where the distance to the obstacle is smaller than the total stopping distance. Brake-and-steer systems attempt to improve this situation by triggering a lane change or a swerve if the system detects that straight line braking is not enough. However, to date and to the best of our knowledge,  these systems either use a simple heuristic(with or without path planning\cite{Choi2012}) or learn a policy using Reinforcement Learning\cite{Chae2017} in a simplified simulation.

In this work we build a recurrent model that predicts the movement of obstacles, pedestrians and vehicles in a way inspired by \cite{worldmodels} and \cite{Chae2017}. Under the hood, our model has the following key components: a Variational Autoencoder (VAE) which reduces the dimensionality of the data and compresses incoming observations into a latent representation, an RNN which learns to predict the next latent representation given the current one and a controller that uses these latent representations and learns to take actions using Deep Deterministic Policy Gradients (DDPG)\cite{DDPG}.  We extend the simulation system of \cite{CARLA} to include both pedestrian and car collisions, along with randomized pedestrians on sidewalks and vehicular traffic on the road around the collision site and focus on Time to Collision (TTC) of 1.5 seconds and under \cite{Chae2017}.  

The major assumptions that we make is that other traffic participants cannot themselves actively avoid the collision by braking or changing direction. We present our results in \ref{sec:results} and show that training a brake-and-steer policy in a complex simulation could reduce the amount of pedestrian and car occupant injury compared to the baseline of braking in a straight line, especially for time-to-collision (TTC)[9] values between 0.5 and 1.5 seconds. In our tests, using standard models of injury as a function of delta-v for vehicle occupants and as a function of impact speed for pedestrians, we see a reduction of up to 60\% in the collision rate for pedestrians and car occupants. 

\begin{figure}[tbp!]
 \includegraphics[width=\columnwidth,trim={0cm 0.7cm 0cm 0cm}]{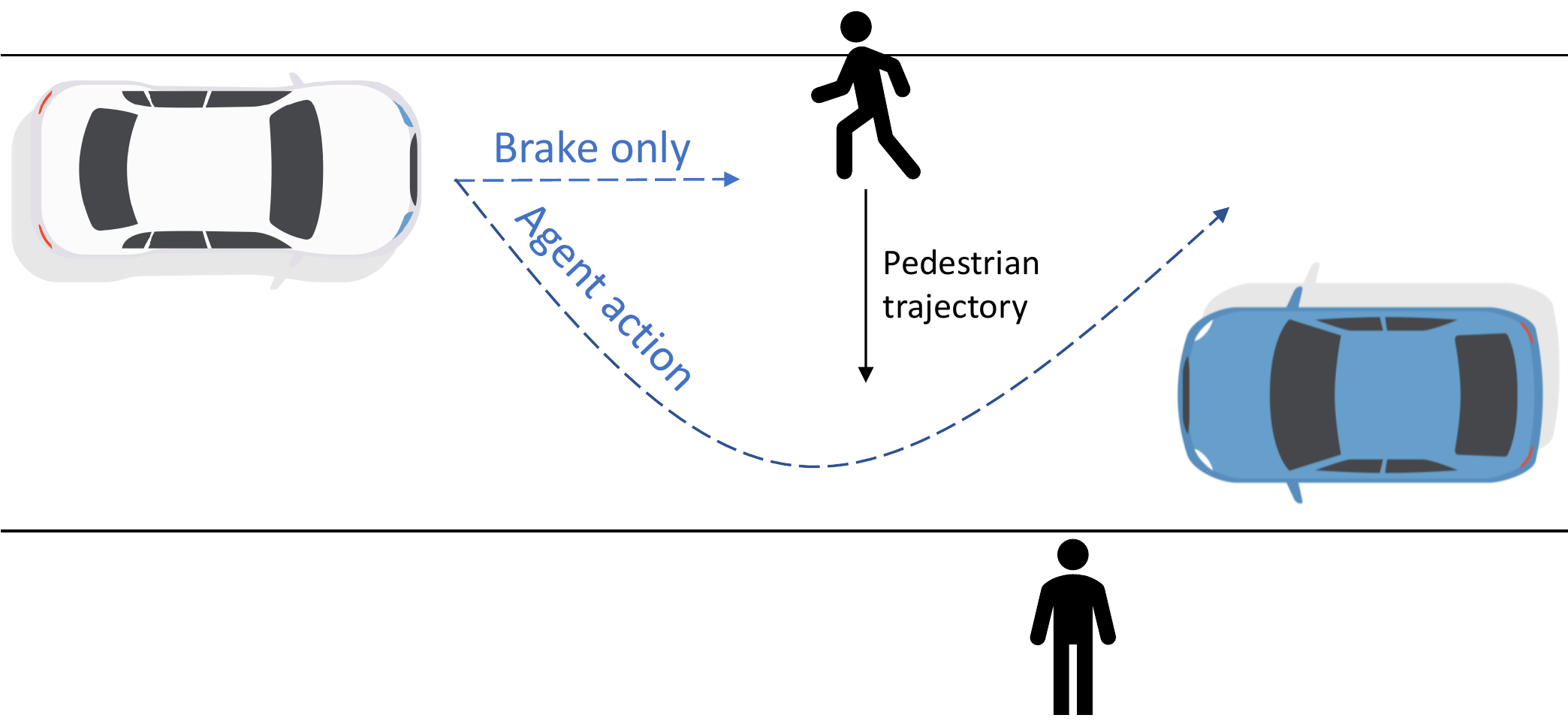}
\caption{The imminent collision mitigation system is trained to choose the action that leads to the least amount of injury, as measured by empirical models which relate delta-v to measures of injury and loss. Our simulated environment contains a mixture of pedestrians and vehicles that can take part in an imminent collision scenario. In this particular example we show a pedestrian that is running across the street in front of the vehicle containing the collision mitigation system, along with incoming traffic and a pedestrian on the sidewalk.}
\label{fig:introfig}
\end{figure}

\section{Related Work}\label{sec:related-work}
In this section we review related work and collision mitigation and on injury risk estimation.
\subsection{Related Work}

In \cite{Kaempchen2009}, the authors present an assessment algorithm that determines if collisions are avoidable by taking into consideration all vehicles detected and their constraints and having the emergency braking being triggered only when collision becomes unavoidable. Similar to our work, their approach does not rely on assumptions regarding vehicle path or road infrastructure, but only vehicle constraints and other detected traffic participants, and their approach can be applied to various crash scenarios, unlike other approaches that only model rear-end collisions.
In an earlier study, \cite{Shiller1998} present a tool for determining the optimal maneuver in collision-avoidance scenarios. A clearance curve is generated for varying speeds, along which hazard states can be avoided only by an optimal maneuver. Their result show that at high speeds, lane-change is more advantageous than a full stop, this highlighting the importance of a more diversified action space. 
In \cite{Ferrara2009} a cruise control system is presented, aimed at reducing collisions within a platoon using a “collision avoidance” mode involving either emergency brake or collision avoidance action whenever new data is acquired from sensors. \cite{Brannstrom2010} propose a model based collision avoidance algorithm in order to approximate a set of actions (break, accelerate or steer) that can be taken by the driver and determine whether immediate assistance is required. Similar to our work, the model can be used for all traffic scenarios and all kinds of traffic participants. \cite{Choi2012} propose a nonlinear model predictive control that aims at lowering the risk of other hazardous situations that result from the vehicle’s attempts to avoid a collision. The model takes into account the vehicle dynamics (minimum and maximum steering wheel angle and acceleration) by implementing a 2-level architecture: a controller that provides the path/state which avoids the collision, and another controller that aids the vehicle at following the proposed path.

A system capable of early detecting a pedestrian’s intention of crossing the road and performing an evasive maneuver if avoidance by braking is impossible is presented in \cite{Köhler2013}. However, they rely on the existence of a Road Side Unit placed in dangerous road spots in order to detect pedestrian intention and send this information to the On Board Unit placed in the vehicle.

As opposed to most previous research, \cite{Chae2017} propose a model-free collision avoidance system using Deep Reinforcement Learning (DRL). They derive a balanced reward function for an autonomous braking system based on DRL, where the action space allows 4 choices: no braking, weak braking, medium and strong. The reward function consists of one component that penalizes the agent for braking too early while the second one is a penalty for collision with the pedestrian and takes into account the velocity of the vehicle in order to reflect the degree of damage.  Due to an unstable learning performance (collisions rarely occuring), the authors use memory replay\cite{replay} in order to remind the agent of collisions, whatever the present policy. The results show that collision rates are 0 for time-to-collision larger or equal to 1.5 seconds. This prompts us to study TTC values below $1.5s$.

\subsection{Injury risk literature}

Various metrics are being employed in research as a measure of crash severity, while this makes for grim reading they do provide widely adopted quantitative, data driven models of accident outcomes.
They include the Acceleration Severity Index (ASI), Occupant Impact Velocity (OIV) and Delta-V. However, \cite{Gabauer2008} show that the former two do not offer significant predictive advantage over the latter. Since its emergence in the 1970s, Delta-V has been the traditional metric for crash severity and is defined as the absolute change between pre-collision velocity and post-collision velocity, with the assumption that larger differences in velocities are correlated with more severe injuries (\cite{Shelby2011}):

\begin{equation} \label{deltav}
    \Delta v = |v_{after} - v_{before}|
\end{equation}

In \cite{Rosen2009} the fatality risk of pedestrians as given by the vehicle's speed on impact using the GIDAS dataset (German In-Depth Accident Study) is studied. The dataset includes data from  2127 pedestrians that were involved in accidents between 1999 and 2007. They present a now widely-adopted approximation of the fatality risk as:
\begin{equation} \label{eq:pedestrian}
    P = \frac{1}{1+e^{6.9 - 0.09|v|}}
\end{equation}
where $v$ is the velocity when impacting the pedestrian (km/h).

The authors of \cite{Joksch1993} introduced a widely used model for the probability of a severe injury of car occupants during a frontal impact, which was later confirmed and extended by \cite{Evans1994} in a larger study. Both papers analyzed crash tests that were published by National Highway Traffic Safety Administration (USA) in order to determine how the risk of severe injury of the front seat occupants of a passenger car is influenced by the impact speed during a frontal crash. They approximate the fatal injury risk to be given by
\begin{equation} \label{eq:occupant}
    P = \Big ( \frac{\Delta v}{71} \Big )^{4}
\end{equation}
where $\Delta v $ is expressed in miles/hour, and the value of $P$ is clamped to $1.0$. Moving forward, and without prejudice, we will be using Equations \ref{eq:pedestrian} and \ref{eq:occupant} as quantitative models of accident severity. Nothing in the coming sections requires these particular models and they could be substituted out at the reader's discretion. What we propose here is a framework rather than a fixed implementation or a judgment call on the correct way to measure accident severity.

\section{Learning to Avoid and Mitigate Collisions}\label{sec:method}
Here we investigate the role of machine learning in managing and mitigating collisions. Our goal is a lower collision cost, especially in the cases of lower time-to-collision values. Our ``agent'' is the vehicle we can execute steering and brake control over.  At runtime our agent is provided only with a sequence of images of the road scene. Given this we need to learn a mapping from images to a 2-channel control sequence which improves outcomes over conventional baseline straight line  braking. In this way the system can choose as how and when to brake and steer as a function of instantaneous speed and time to collision. We will consider and compare two loss/reward strategies which will be explained in Section \ref{sec:Reward}.
In the remaining subsections we assume some familiarity with the recent reinforcement learning literature in which for completeness we give  the details of our learning procedure. However the experiment and results in Sections \ref{sec:experimental-setup} and \ref{sec:results} can be consumed / read as independently of this section.

\subsection{Reinforcement learning using DDPG}\label{train:DDPG}

We use Deep Deterministic Policy Gradient (DDPG)\cite{DDPG} because it is off-policy, making it suitable for rapidly discovering optimal policies in simulation by employing a stochastic policy for improved exploration, while tackling the much easier task of learning a deterministic policy. DDPG employs two networks, a policy network (the actor) and a Q-value estimator network (the critic). Using the current state as as input to both networks, the actor will output actions from a continuous action space, while the critic estimates a Q value based on the output of the actor. The actor network weights are then updated using deterministic policy gradient\cite{DPG}, and the critic weights are updated using the gradient of the temporal-difference signal, similar to \cite{DPG}. 

DDPG uses experience replay\cite{replay} as a variance-reduction technique and target networks\cite{DPG} to stabilize training. As a difference from DPG\cite{DPG}, rather than directly copying the weights of the local networks $\theta_{L}$ to the target networks $\theta_{T}$ every N steps, DDPG performs a soft update of its target network weights $\theta_{T}$ using an update rate $\tau$ between $0$ and $1$:
\begin{equation}
    \theta_{T} = (1 - \tau) * \theta_{T} + \tau * \theta_{L}
\end{equation}

The local critic network is updated by minimizing the loss function:
\begin{equation}
    \mathcal{L}({\mathrm{\theta^{Q}_{L}}})= \frac{1}{N}\sum\limits_{i} \Big((R+\gamma Q(s_{i+1}, a_{i+1}|\theta^{Q}_{L})) - Q(s_i, a_i|\theta^{Q}_{T})\Big)^2
\end{equation}

Here, $R$ is the reward, $Q$ is the Q-value approximator function, the local critic network is parametrized by $\theta^{Q}_{L}$, the target critic network is parametrized by $\theta^{Q}_{T}$, $\gamma$ is the time-horizon discount factor and $N$ is the size of the minibatch sampled from the experience replay buffer. The current state and action are represented by $s_i$ and $a_i$, and the future state and action are represented by $s_{i+1}$ and $a_{i+1}$.
The local actor network is updated using the gradient of the local critic network with respect to the actions $a_i$, multiplied by the gradient of the local actor network with respect to $\theta^{\mu}_L$ (chain rule) \cite{DPG}:
\begin{equation}
    \nabla_{\theta^{\mu}_L} \mu = \frac{1}{N}\sum\limits_{i} \big [ \nabla_{a} Q(s_i, a_i|\theta^{Q}_L) \nabla_{\theta^{\mu}} \mu(s_i|\theta^{\mu}_L) \big ]
\end{equation}

Here, $\mu$ is the policy function, the local actor network is parametrized by $\theta^{\mu}_{L}$ and the target actor network is parametrized by $\theta^{\mu}_{T}$. The parameterisation, origin, and form of the reward $R$, action $a$, and state $s$ will now be explained in the following subsections.

\subsection{Reward structure}\label{sec:Reward}
As this research aims to be an objective comparison of the agent's behaviour as a result of the learning strategies, we propose two reward structures for the reinforcement-learning agent: one where the penalties are uniform (-1 for any type of collision) and another one that follows the literature on severe risk injury and pedestrian fatality risk.

While the work of \cite{Chae2017} focuses on pedestrians only, our proposed reward structures will also take into account another category of traffic participants: car-occupants. Secondly, as injury risk has been approximated in previous research, we hope that the second reward structure will more closely follow real collision scenarios, where car-occupants are better protected than pedestrians due to the car's energy absorption properties. Additionally, this empirical approach allows us to avoid any hyper-parameters or weights in the reward functions.

\vspace{3mm}{\textbf{Reward Strategy 1}} The first reward function is defined as:
\begin{equation}
    R_{1} = -(nr_{ped}+nr_{occ})
\end{equation}
Here $nr_{ped}$ represents the number of pedestrians and $nr_{occ}$  represents the number of car occupants involved in the accident. This strategy simply counts the number of people involved in a collision. 

\vspace{3mm}{\textbf{Reward Strategy 2}} The second reward function combines the empirical models of ``injury'' Equations (\ref{eq:pedestrian}) and (\ref{eq:occupant}) and so explicitly accounts for the degree of injury, and is defined as:
\begin{equation}
    R_{2} = -\Big( nr_{ped}\frac{1}{1+e^{6.9-0.09|v|}} + nr_{occ} \Big (\frac{\Delta v * 0.621}{71} \Big)^{4} \Big)
\end{equation}

In the occupant injury component, $0.621$ is used to convert the speed from km/h in miles/hour to fit Joksch's\cite{Joksch1993} original model.\vspace{3mm}

Note that when it comes to comparing outcomes using Strategy 1 or Strategy we will use, in both cases,  the empirically derived Equations \ref{eq:pedestrian} and \ref{eq:occupant} to numerically and continuously represent total injury.

\subsection{Action space}\label{actionspace}
In line with previous model-based research that includes steering and braking as collision-avoidance options, we propose an extension of \cite{Chae2017} agent's action space to include both braking and steering. Additionally, we let the actions be continuous instead of discrete in order to allow more degree of control. 

The actions $a_t$ are encoded as a pair of integer values between -1 and 1, the first value encoding the steering angle, the second value encoding the position of the throttle(for positive values) or brake(for negative values). For steering, the range $-1$ to $1$ maps the range between the maximum left and the maximum right steering angles. For braking, the range $-1$ to $0$ maps a braking force causing a decelleration between $-9.8m/s^2$ to $0m/s^2$. 

\subsection{General Simulation Architecture}\label{generalarch}

\begin{figure}[t!]
 \includegraphics[width=\columnwidth,trim={0cm 0cm 0cm 0cm}]{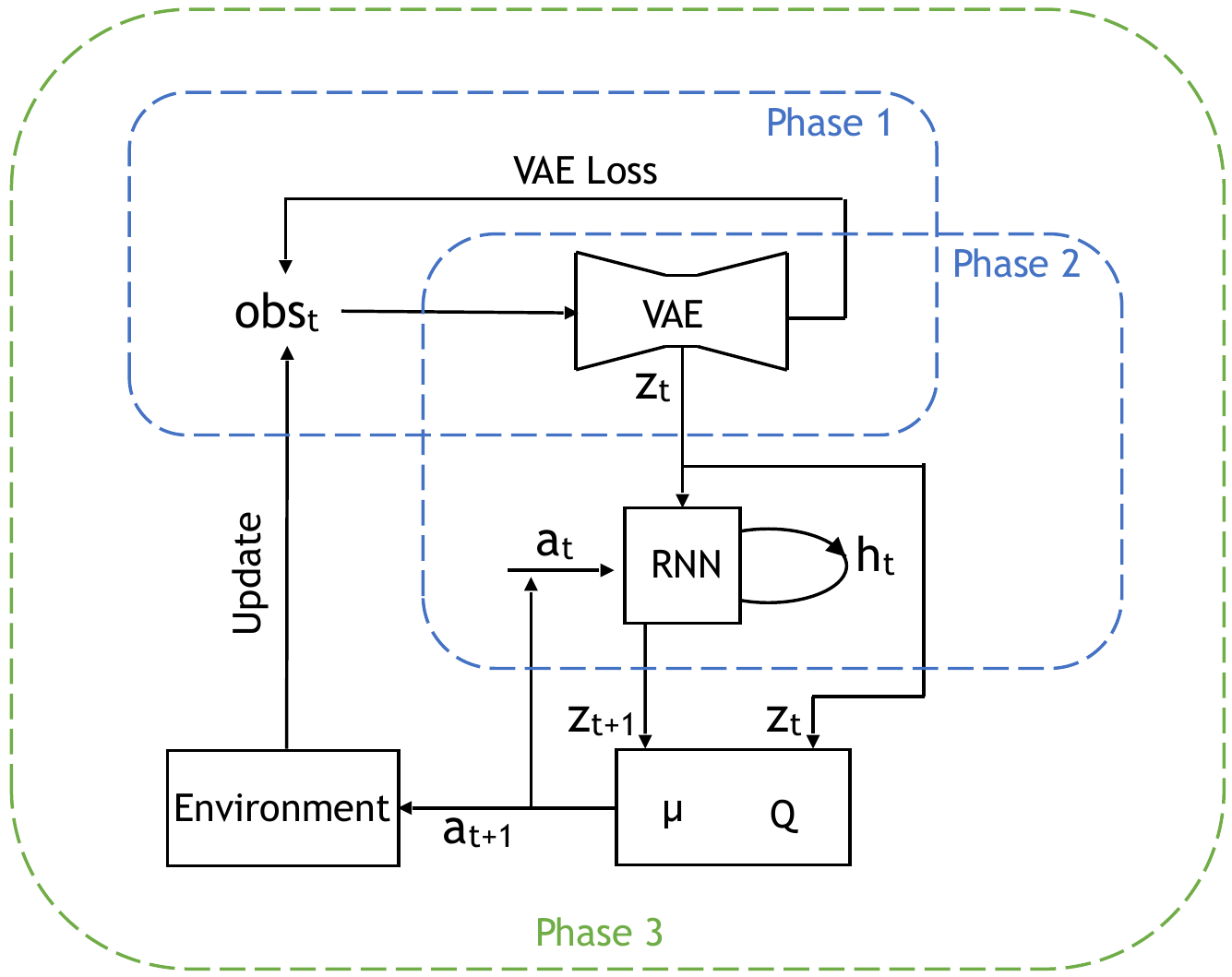}
\caption{An overview of our architecture. The dashed rounded rectangles encompass the components used in each phase, as described in \ref{generalarch} and the following subsections.}
\label{fig:architecture}
\end{figure}

\begin{figure}[t!]
 \includegraphics[width=\columnwidth,trim={0cm 0cm 0cm 0cm}]{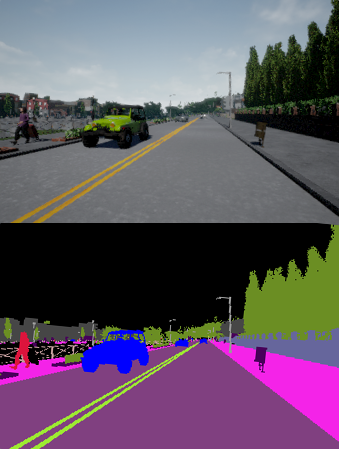}
\caption{Images from the CARLA environment. On the top we show an RGB image depicting a street view with traffic. On the bottom we show the semantic segmentation of the top image, with cars in blue and pedestrians in red.}
\label{fig:CARLA}
\end{figure}

Clearly in this context the use of simulation is required. One cannot readily or ethically explore the space of accident vs. action in real life. Accordingly, our system as presented here appears as a back-end used in conjunction with CARLA's semantic segmentation output based on Citiscapes\cite{citiscapes} classes,which is fed to our system in the shape of 64*64 1-channel images, with 13 corresponding classes(None, Buildings, Fences, Other, Pedestrians, Poles, RoadLines, Roads, Sidewalks, Vegetation, Vehicles, Walls, Traffic Signs) encoded as integer values between 0-12. Our choice is based on the recent performance of real-time semantic segmentation approaches\cite{chen2017rethink},\cite{largekernel}, and we assume that such a system is now readily available as an input pre-processor.
Similar to \cite{worldmodels}, we develop and train our system in multiple phases, as this allows us to use supervised learning where possible and to test individual components. We begin by recording observations $obs_t$ and actions $a_t$ produced by an agent exploring the environment of CARLA using the built-in waypoint based autopilot\cite{CARLA} with Gaussian noise added to the controls. The observations $obs_t$ that we collect are represented by tuples of a semantically-segmented(using Citiscapes classes) images of the environment $I_t$, as shown in Fig. \ref{fig:CARLA} and the vehicle forward velocity $v_t$, for each time $t$. The actions $a_t$ are encoded as described in section \ref{actionspace}.
In the first phase, a Variational Autoencoder (VAE)\cite{Kingma2013} learns to compress the incoming observations from the environment $obs_t$ into a latent representation $z_t$ with a Gaussian distribution. In the second phase, we train a Recurrent Neural Network (RNN) to predict the next latent representation $z_{t+1}$, given the current latent representation $z_t$ and the current action $a_t$. Finally,in the third phase, we train the agent using DDPG by encoding the current incoming frame using the VAE, predicting the next state using the recurrent network and using the latent representation $z_t$ and the predicted state $z_{t+1}$ as input to our controller, giving the agent access to the current state of the environment and to a prediction of the future state. The agent then takes an action in the simulated Carla environment. An overview of the architecture of the system is shown in Fig. \ref{fig:architecture}.

\subsection{Phase 1}
Following \cite{worldmodels}, in Phase 1 we train a Variational Autoencoder (VAE)\cite{Kingma2013} to compress incoming 2D observations (images) $obs_{t}$ into a 128-length latent representation $z_{t}$ with a Gaussian distribution. The literature on VAEs is vast, and the reader is invited to study, for example, \cite{Rezende2014} and \cite{Kingma2013} for a more in-depth look at VAEs. 
The main reasoning behind using a VAE is two-fold: 1)we wish to reduce the dimensionality of our data such that it becomes easier to train the RNN and controller and 2)we wish to only keep features that are relevant to encoding the position and type of objects found in the incoming semantically-segmented images.

\subsection{Phase 2}
\begin{figure}[t!]
 \includegraphics[width=\columnwidth,trim={0cm 0cm 0cm 0cm}]{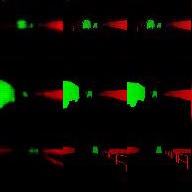}
\caption{Images obtained by applying the VAE decoder on latent representations. The right column represents decoded state $z_{t-1}$. The middle column represents decoded state $z_t$. The left column represents the output of the decoder applied to state $z_{t+1}$ as predicted by the RNN using $z_t$ and the internal state of the LSTM cell. The green blob represents a semantically-segmented vehicle coming towards the camera, while the red patches represent fences on the side of the road.}
\label{fig:latent_decoder}
\end{figure}

In phase 2, we first convert all of the recorded 2D image observations $obs_t$ into latent space representations $z_t$. Following this we produce training data by creating a sequence of tuples $(z_t, z_{t+1}, a_t)$ for $t$ between 1 and $N$, the total number of recorded frames in a sequence. 
We then train an LSTM-RNN\cite{LSTM} to predict the next latent representation $z_{t+1}$  given the current latent representation $z_t$ and the current action $a_t$ using the sequence of tuples. An example of applying the VAE decoder on the predicted latent representations can be seen in Fig \ref{fig:latent_decoder}.
The RNN layout uses an LSTM cell with 512 hidden units. Details about the implementation can be referenced from \cite{worldmodels}.

\subsection{Phase 3}
In phase 3, we train our agent in the CARLA environment using DDPG, as described in \ref{train:DDPG}. For each incoming observation $obs_t$, we transform it using the VAE into a latent representation $z_t$ and predict a future state $z_{t+1}$. The input to both the policy and the Q-value functions is a concatenation of $z_t$ and $z_{t+1}$, giving the agent access to the current state of the environment and to a prediction of the future state.

\section{Experimental Setup}\label{sec:experimental-setup}

\subsection{Generating and Simulating Collision Scenarios}

To generate a diverse set of imminent collision scenarios, we randomly sample collision parameters from the following distributions:
\begin{itemize}
    \item $TTC \sim \mathcal{U}(0.25,\,1.5) s$
    \item Collision velocity $v_c \sim \mathcal{U}(1,\,30) m/s$
    \item Pedestrian velocity $v_p \sim \mathcal{U}(1,\,5) m/s$
    \item Number of cars $ n_c\sim [0...10]$
    \item Number of other pedestrians $ n_p\sim [0...10]$
    \item Probability of pedestrian infraction $ P_p\sim \mathcal{U}(0,\,1)$
    \item Probability of car infraction $ P_c\sim \mathcal{U}(0,\,1)$
\end{itemize}

For each scenario, if the probability of pedestrian collision is higher than $0.5$, we generate a random trajectory that intersects the pedestrian with the car at a point that will be reached in $TTC$ seconds if the velocity of the car is maintained. Similarly, if the probability of a car collision is higher than 0.5, we spawn a vehicle that will intersect with the car at a point that is $TTC$ seconds away if the velocity is kept constant.

\subsection{Training}

We train the VAE for 2 epochs on a dataset containing $200000$ frames. We train the RNN for 10 epochs on 850 sequences of approximately 230 frames each, captured in the CARLA environment at 15FPS, with a time-limit of 15 seconds for each sequence. 
We make use of the Adam \cite{Kingma2014} solver for all training, with an initial learning rate of 0.0001, and a batch size of 32. No other hyperparameters are set or used.

\subsection{Testing}

To test the effectiveness of our policies, we create a bank of test scenarios by sweeping through TTC, collision speeds and traffic participant setups, with fixed seeds for any randomly-generated component to allow for repeatability. We then test the 2 policies and the baseline policy using identical test scenarios, and record for each collision scenario, along with the scenario setup, the following metrics:
\begin{itemize}
    \item Whether there was a collision with a pedestrian
    \item Whether there was a collision with a vehicle
    \item Whether there was a collision with a static object
    \item Injury severity for each collision, for each participant
    \item Velocity of car at the beginning of the collision
    \item Whether the car has left its lane
\end{itemize}

These metrics allow us to study,for each policy type and for each Time-To-Collision interval:
\begin{itemize}
    \item The percentage of pedestrian collisions avoided
    \item The percentage of car collisions avoided
    \item The percentage of collisions with static obstacles (e.g. poles, walls etc)
    \item The percentage of severe injuries
\end{itemize}

\subsection{Performance}
The agent network takes approximately 100 ms for images with a resolution of $64\times64$ on an Intel I7 processor and 30 ms for the same resolution on an NVIDIA Tesla GPU.

\section{Results}\label{sec:results}

\begin{figure*}[h!]
 \includegraphics[width=\textwidth,trim={0cm 0cm 0cm 0cm}]{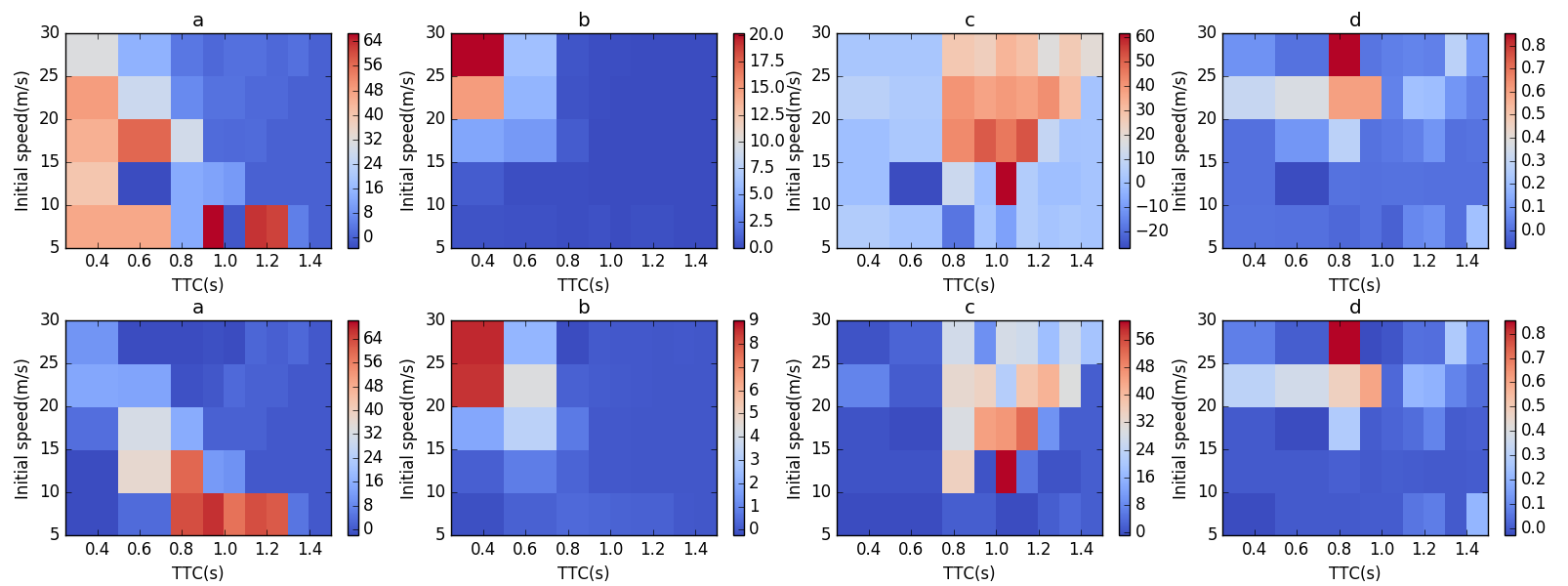}
\caption{First row presents results for the policy trained using Reward 1. Second row presents results for the policy trained using Reward 2. Column descriptions: a) Pedestrian collision rate percentage point reduction over baseline; b) Pedestrian severe injury risk percentage point reduction over baseline; c) Car collision rate percentage point reduction over baseline; d) Car occupant severe injury risk percentage point reduction over baseline; The more red an area is(i.e, the higher the numbers), the lower the rate of collision or risk of severe injury compared to a baseline of straight-line braking-only.}
\label{fig:crash_avoid}
\end{figure*}

\begin{table*}[!h]
\begin{center}
\begin{tabular}{@{} |c|*{12}{C{0.7cm}} @{}}
  \hline
   \multirow{3}*{\bf Policy type} & \multicolumn{3}{c|}{\bf TTC 0.5s}& \multicolumn{3}{c|}{\bf TTC 0.75s}& \multicolumn{3}{c|}{\bf TTC 1.0s}& \multicolumn{3}{c|}{\bf TTC 1.5s} \\
   & 10 (m/s) &20 (m/s)&\multicolumn{1}{C{0.7cm}|}{30 (m/s)}
   & 10 (m/s) &20 (m/s)&\multicolumn{1}{C{0.7cm}|}{30 (m/s)}
   & 10 (m/s) &20 (m/s)&\multicolumn{1}{C{0.7cm}|}{30 (m/s)}
   & 10 (m/s) &20 (m/s)&\multicolumn{1}{C{0.7cm}|}{30 (m/s)} \\ \hline
   Straight-line braking & 0.886&5.40&\multicolumn{1}{c|}{6.08}&0.263&0.360&\multicolumn{1}{c|}{1.650}&0.015&0.064&\multicolumn{1}{c|}{0.828}&0.0&0.0&\multicolumn{1}{c|}{0.0} \\ \hline
   Brake\&Steer, reward 1 & 0.168&2.462&\multicolumn{1}{c|}{2.860}&0.0&0.189&\multicolumn{1}{c|}{1.799}&0.0&0.0&\multicolumn{1}{c|}{0.0}&0.0&0.0&\multicolumn{1}{c|}{0.0} \\ \hline
   Brake\&Steer, reward 2  & 0.054&1.087&\multicolumn{1}{c|}{3.487}&0.0&0.172&\multicolumn{1}{c|}{1.633}&0.0&0.0&\multicolumn{1}{c|}{0.0}&0.0&0.0&\multicolumn{1}{c|}{0.0} \\ \hline
\end{tabular}
\end{center}
\caption{Pedestrian probability (\%) of severe injury}
\label{tab:DeviationResults1}
\vspace{-0.5cm}

\end{table*}

\begin{table*}[!h]
\begin{center}
\begin{tabular}{@{} |c|*{12}{C{0.7cm}} @{}}
  \hline
   \multirow{3}*{\bf Policy type} & \multicolumn{3}{c|}{\bf TTC 0.5s}& \multicolumn{3}{c|}{\bf TTC 0.75s}& \multicolumn{3}{c|}{\bf TTC 1.0s}& \multicolumn{3}{c|}{\bf TTC 1.5s} \\
   & 10 (m/s) &20 (m/s)&\multicolumn{1}{C{0.7cm}|}{30 (m/s)}
   & 10 (m/s) &20 (m/s)&\multicolumn{1}{C{0.7cm}|}{30 (m/s)}
   & 10 (m/s) &20 (m/s)&\multicolumn{1}{C{0.7cm}|}{30 (m/s)}
   & 10 (m/s) &20 (m/s)&\multicolumn{1}{C{0.7cm}|}{30 (m/s)} \\ \hline
   Straight-line braking & 0.0&0.372&\multicolumn{1}{c|}{0.829}&0.008&0.603&\multicolumn{1}{c|}{1.240}&0.006&0.042&\multicolumn{1}{c|}{0.015}&0.001&0.002&\multicolumn{1}{c|}{0.245} \\ \hline
   Brake\&Steer, reward 1& 0.001&0.467&\multicolumn{1}{c|}{0.579}&0.001&0.754&\multicolumn{1}{c|}{0.495}&0.0&0.052&\multicolumn{1}{c|}{0.001}&0.0&0.0&\multicolumn{1}{c|}{0.0} \\ \hline
   Brake\&Steer, reward 2& 0.0&0.001&\multicolumn{1}{c|}{0.663}&0.0&0.124&\multicolumn{1}{c|}{0.0}&0.0&0.0158&\multicolumn{1}{c|}{0.0}&0.0&0.0&\multicolumn{1}{c|}{0.0} \\ \hline
\end{tabular}
\end{center}
\caption{Car occupant probability (\%) of severe injury}
\label{tab:DeviationResults2}
\vspace{-0.5cm}

\end{table*}

\begin{table}[!h]
\centering
\begin{tabular}{|l|l|l|l|l|l|l|}
\hline
TTC(s) &0.5& 0.75 & 0.9 & 1.1 & 1.3 & 1.5 \\ \hline
Col. rate \% (ours)       &\textbf{21.8} &\textbf{5.0} &\textbf{1.4} &\textbf{0.0} &\textbf{0.0} &0.0 \\ \hline
Col. rate \%  \cite{Chae2017} &NA &NA & 61.29 & 18.85& 0.74 & 0.0\\ \hline
\end{tabular}
\caption{Pedestrian collision rate (\%) comparison between our results and those of Chae et al.\cite{Chae2017}.}
\label{ref:ped_comp}
\end{table}

Fig. \ref{fig:crash_avoid} shows, on each column, in the following order: a)Pedestrian collision rate percentage point improvement over baseline; b)pedestrian severe injury risk percentage point improvement over baseline; c)car collision rate percentage point improvement over baseline; d)car occupant severe injury risk percentage point improvement over baseline; All are expressed as a function of TTC and initial velocity. First row presents results for the policy trained using Reward $R1$, while the second row presents results for the policy trained using Reward $R2$. We notice that both policies offer very favourable outcomes as compared to the baseline of braking-only. 

We observe a significant increase in the number of collisions avoided as compared to baseline, especially for TTC values between $0.25s$ and $1.0s$ for pedestrians, and for TTC values between $0.75s$ and $1.3s$ for car collisions. We observe no overall degradation of performance as compared to the braking-only baseline, suggesting that the controller has learned to effectively gauge when simple braking is enough and when steering input is needed to mitigate the effects of a collision. 

We also observe a significant improvement in the risk of severe injury for pedestrians, especially in the problematic\cite{Chae2017} area of TTC under $0.9s$ and speeds over $15m/s$.

Table \ref{tab:DeviationResults1} compares the performance of the trained agent against the braking-only baseline for pedestrian risk of severe injuries. Both policies lead to a lower incidence of severe injuries compared to straight-line braking, with the policy trained using $R2$ showing outright improvement in all areas, with the exception of TTC $0.75s$ at $30m/s$.

Table \ref{tab:DeviationResults2} compares the performance of the trained agent against the braking-only baseline for car occupant injuries. We observe that the policy trained using $R2$ shows an outright improvement in all areas.

Table \ref{ref:ped_comp} compares the pedestrian collision results obtained using our $R2$ policy with the results published by \cite{Chae2017}. We show a reduction in the rate of severe injuries across the whole TTC range between $0.9s$ and $1.5s$, and show much lower collision rates at TTC of $0.75s$ and $0.5s$ than the rates reported by \cite{Chae2017} for TTC of $0.9s$.

\section{Conclusions}\label{sec:conclusions}

This paper has asked an unusual question: ``what might an autonomous vehicle do during an accident, including the cases where a collision is inevitable?''. There is, of course, no obviously correct answer, nor do we intend here to offer a view on what \emph{should} happen. Instead, we ask what control behaviours are learned if we utilize existing models of injury severity and at the same time leverage the ability of an Autonomous Vehicle to control its complete trajectory  - thereby differentiating from conventional ADAS systems.  
 
Our architecture makes several contributions to the topic of investigating collision mitigation using simulation and machine learning. The training of the VAE distills relevant information from the CARLA simulations such that both the RNN and the controller become much easier to train. And of course the use of Reinforcement Learning in the controller allows the system freedom to learn complex control profiles.

We compare our outcomes, which allow for steering and braking control, to those arising  from straight-line emergency braking systems. We trained our system using two reward constructs - one which simply counts participants in collisions, and one which takes into account empirically derived measures of severity. In all three cases (our two strategies and the baseline) we measure the outcome using the same empirical measures of injury, Eq.(\ref{eq:pedestrian}) and Eq.(\ref{eq:occupant}).

We have presented a vision-based emergency collision mitigation system that, under simulation, reduces the total severity of human injury. 
Our empirical results show that allowing the vehicle to brake \emph{and} steer yields preferable outcomes when used with the policies we learned. In some cases by a 60\% margin. 

\section{Acknowledgements}\label{sec:acknowledgements}

This work was supported by an Oxford-Google DeepMind Graduate Scholarship and EPSRC Programme Grant EP/M019918/1. The authors would also like to thank Valentina Musat and Tom Bruls for their help.

\bibliographystyle{IEEEtran}
\bibliography{ITSC}

\end{document}